# AN EFFECTIVE SPOKEN LANGUAGE UNDERSTANDING APPROACH LEVERAGING QUERY-ADAPTIVE INTENT DETECTION

*[1]†Bing-Long Kuo, [1]†Yu-Hsuan Fang, [1]Tien-Hong Lo, [2]Hsueh-Wei Liao, and [1]Berlin Chen*

**[1]National Taiwan Normal University, Taipei, Taiwan**
**[2]Realtek Semiconductor Corp., Taiwan**
†These authors contributed equally to this work.
E-mail: {614k0022c, andyfang, teinhonglo, berlin}@ntnu.edu.tw; liao_hsueh_wei@realtek.com

## ABSTRACT

Large language models enable fine-tuning-free spoken language understanding (SLU) through in-context learning, but direct query-based retrieval may select semantically similar yet intent-irrelevant demonstrations. This paper presents Query-Adaptive Intent Detection via Prototype Extraction (APEX), a two-stage retrieval framework that first selects candidate intents using intent-level prototypes and then retrieves contextual exemplars within the selected intent space. Experiments on the SLURP dataset evaluate joint SLU performance using Exact Match Accuracy (EMA) under full-transcript and perplexity-stratified conditions, with additional retrieval analysis on ASR-generated transcripts. Compared with query-based retrieval, APEX improves EMA by 4.34 and 3.97 percentage points with Qwen3-4B and Gemma-3-4B, respectively, on full transcripts. Under high-perplexity inputs, APEX also yields EMA gains of 7.66 and 2.13 percentage points. Retrieval analysis further indicates that prototype-guided candidate selection achieves higher ground-truth intent coverage across multiple input conditions.

***Keywords:*** *Spoken language understanding, large language models, in-context learning, prototype-guided retrieval, intent detection.*

## 1. INTRODUCTION

Spoken language understanding (SLU) aims to identify user intents and extract semantic slots from speech inputs and serves as an essential component in voice assistants and speech-interactive systems [1,2]. Recent advances in large language models (LLMs) further enable semantic prediction through prompting and in-context learning (ICL) without task-specific fine-tuning [3–6]. Compared with conventional supervised settings, fine-tuning-free inference can reduce the need for additional task-specific parameter updates, making it attractive when retraining or adaptation is costly.

Fine-tuning-free SLU may still be sensitive to variations in input quality and linguistic form. ASR transcription errors can distort semantic cues and affect downstream SLU performance [7-10], while linguistically less predictable utterances may introduce additional uncertainty during semantic interpretation. These effects become particularly relevant when intent categories share similar expressions or semantic patterns, since small changes in the input may alter the evidence available to the downstream prediction process.

Retrieval-based prompting provides contextual demonstrations that can support LLM inference without task-specific parameter updates [6,11]. In conventional query-based retrieval, demonstrations are typically ranked according to their similarity to the input query. However, local semantic similarity does not always imply intent consistency, especially when utterances from different intents contain overlapping expressions. To address this issue, we introduce Query-Adaptive Intent Detection via Prototype Extraction (APEX), which incorporates class-level intent information into the demonstration retrieval process.

Experiments are conducted on the SLURP [2] dataset. Final SLU performance is evaluated using Exact Match Accuracy (EMA) under full-transcript and perplexity-stratified conditions, while ASR-generated transcripts from Whisper Large-v2 and Whisper Tiny-en [8] are used separately to examine retrieval behavior under transcription noise. Perplexity (PPL) is adopted as a proxy for utterance predictability to divide the test set into Low-PPL and High-PPL subsets. The main contributions of this work are threefold: (1) we present APEX, a prototype-guided two-stage retrieval framework that introduces intent-level candidate selection before contextual exemplar retrieval for fine-tuning-free SLU; (2) we evaluate joint semantic prediction under full-transcript and PPL-stratified conditions; and (3) we analyze retrieval behavior under ASR-generated transcripts and PPL-based subsets to examine how prototype-guided candidate selection affects ground-truth intent coverage.

## 2. RELATED WORK

Recent studies have explored large language models (LLMs) for spoken language understanding (SLU) under zero-shot and in-context learning settings, reducing the need for task-specific parameter updates [5,12,13]. In-context learning can incorporate labeled demonstrations into the prompt to provide task-specific semantic references without additional fine-tuning [6]. These approaches suggest that LLMs can support intent and slot prediction in fine-tuning-free settings, although their performance may depend on how contextual examples are selected and presented.

Retrieval-based prompting further extends in-context learning by selecting demonstrations from a labeled example pool according to their relevance to the input [6,11]. Compared with using a fixed set of demonstrations, retrieval allows the prompt context to adapt to individual queries. However, common example-level retrieval strategies primarily rely on similarity between the query and individual training instances. When semantically overlapping expressions occur across different intents, highly similar examples may not always provide intent-consistent demonstrations, which can introduce ambiguity during subsequent semantic prediction.

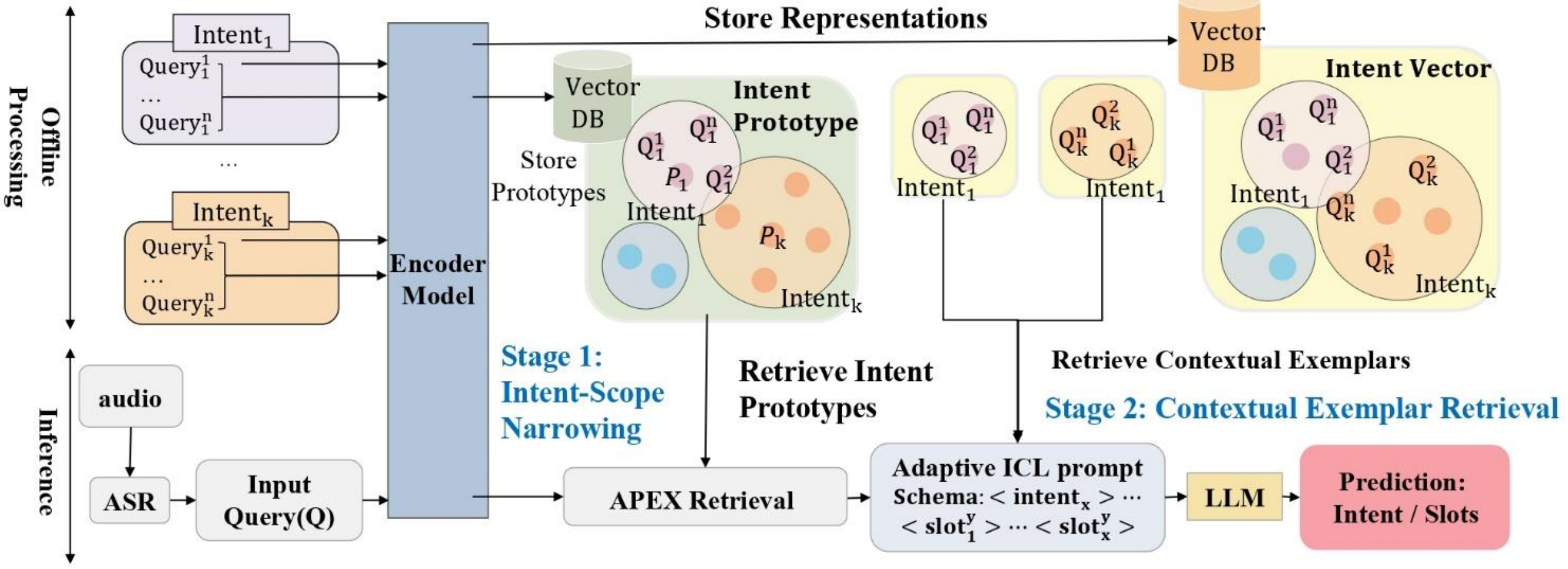

Fig. 1. Overview of the proposed APEX framework. The ASR path is considered only in the retrieval-stage analysis; the main SLU evaluation uses reference transcripts.

Prototype-based representations summarize class-level information using representative embeddings and have been widely studied in few-shot learning [14], including applications to intent detection and joint intent-slot modeling [15,16]. Existing studies typically incorporate prototypes into learned metric spaces, classification objectives, or representation-learning frameworks [14–16]. In particular, prior coarse-to-fine prototype learning focuses on improving few-shot intent classification through learned prototype representations [16]. In contrast, APEX does not introduce an additional prototype-learning objective; instead, it uses fixed intent prototypes as a retrieval prior to narrow the candidate intent space before query-specific exemplar retrieval. This design shifts the role of prototypes from learned decision representations to a lightweight mechanism for organizing in-context demonstrations. The detailed retrieval procedure is presented in Section 3.

## 3. PROPOSED APPROACH

### 3.1. Overview

Query-Adaptive Intent Detection via Prototype Extraction (APEX) is a two-stage retrieval framework for fine-tuning-free SLU. Given an input transcript, APEX first performs intent-level candidate selection and then conducts contextual exemplar retrieval within the resulting candidate space. The retrieved exemplars are incorporated into an in-context prompt for joint intent and slot prediction. This separation allows class-level information to guide the scope of subsequent example-level retrieval. An overview of the framework is illustrated in Fig. 1.

### 3.2. Intent prototype construction

Let f(·) denote the pretrained sentence encoder used to map an utterance into a dense semantic representation. For each intent $i$, APEX constructs an intent prototype by averaging the embeddings of all training utterances associated with that intent:

$$P_i = \frac{1}{N_i}\sum_{j=1}^{N_i} f\left(x_j^{(i)}\right), \tag{1}$$

where $N_i$ denotes the number of training utterances associated with intent $i$, $x_j^{(i)}$ denotes the $j$ -th training utterance of that intent, and $P_i$ denotes the resulting intent prototype. The prototype is constructed once from the labeled training set and remains fixed during inference. This averaging operation provides a compact class-level summary without introducing an additional trainable prototype module. The resulting intent prototypes and individual training-utterance embeddings are stored in a vector database for subsequent candidate intent selection and exemplar retrieval, respectively.

### 3.3. Prototype-guided candidate intent selection

Given an input query $q$, APEX first obtains its semantic representation $f(q)$ and computes the cosine similarity between the query representation and each intent prototype:

$$\text{sim}(q, P_i) = \frac{f(q)\cdot P_i}{||f(q)||\,||P_i||}. \tag{2}$$

The intent prototypes are ranked according to their similarity scores, and the Top-$K$ intents form a query-specific candidate set $C_K(q)$:

$$C_K(q) = \text{TopK}_{i\in I}\,\text{sim}(q, P_i), \tag{3}$$

where $I$ denotes the complete set of intents. The value of $K$ controls the size of the candidate intent space and, consequently, the amount of class-level filtering applied before exemplar retrieval. We therefore examine $K\in\{3,5,10\}$ through the Coverage@K analysis described in Section 4.3.

### 3.4. Contextual exemplar retrieval

After obtaining $C_K(q)$, APEX restricts the retrieval pool to training utterances whose intent labels belong to the selected candidate set. Query-to-example cosine similarity is then computed within this restricted pool, and the top-ranked exemplars are selected as contextual demonstrations. The prompt consists of a system instruction specifying the structured semantic output format, the retrieved input-output demonstrations, and the test query; each demonstration pairs an input utterance with its intent-and-slot annotation. The resulting prompt is provided to the LLM for joint intent and slot prediction. No task-specific parameter update is performed during this process.

For comparison, the query-based retrieval baseline retrieves exemplars directly from the complete training set without intent-level candidate selection. APEX and the query-based baseline otherwise use the same sentence encoder, backbone LLM, and downstream prediction procedure, as detailed in Section 4.2. This comparison is used to examine the effect of introducing prototype-guided candidate filtering before example-level retrieval.

## 4. EXPERIMENTAL SETUP

### 4.1. Dataset and input conditions

We conduct experiments on the SLURP dataset [2], a spoken language understanding benchmark containing natural speech utterances annotated with intents and semantic slots. We use the official data split and evaluate the proposed approach on the test set. Three input conditions are considered. First, reference transcripts are used as the primary input for evaluating final SLU performance. Second, the test utterances are divided into Low-PPL and High-PPL subsets to examine performance under different levels of utterance predictability.

Perplexity is computed using DistilGPT-2 as the language model. Given a token sequence $w_1, \ldots, w_T$, its perplexity is defined as:

$$\mathrm{PPL}(w_{1:T}) = \exp\left(-\frac{1}{T}\sum_{t=1}^{T}\log p(w_t|w_{<t})\right), \qquad (4)$$

where $T$ denotes the number of tokens in the sequence and $p(w_t|w_{<t})$ is the conditional probability of token $w_t$ given its preceding tokens. Higher PPL indicates that an utterance is less predictable under the language model. Following Tukey's 1.5×interquartile range (IQR) criterion [17], utterances with valid PPL scores above the resulting upper threshold are assigned to the High-PPL subset, while the remaining valid-PPL utterances form the Low-PPL subset. Valid PPL scores are obtained for 2,952 of the 2,974 test utterances, yielding 2,482 Low-PPL (84.1%) and 470 High-PPL (15.9%) samples. For our fine-tuning-free experiments, the full-set results are evaluated on all 2,974 test utterances, whereas the PPL-stratified results are reported on the 2,952 utterances with valid PPL scores. Finally, transcripts generated by Whisper Large-v2 [8] and Whisper Tiny-en [8] are used separately to examine retrieval behavior under ASR transcription noise. These ASR-generated transcripts are used only for retrieval-stage analysis and not for reporting final downstream SLU performance.

### 4.2. Models and retrieval settings

We evaluate APEX using Qwen3-4B [18] and Gemma-3-4B [19] as the backbone LLMs. The pretrained BAAI/bge-large-en-v1.5 sentence encoder [20] is used for both prototype construction and query-to-example retrieval. Intent prototypes are computed from the training set as described in Section 3.2 and remain fixed during inference. For all downstream APEX experiments reported in Table 1, we use K = 5 candidate intents. Both APEX and the query-based retrieval baseline use a retrieval budget of five contextual exemplars for in-context prediction. For retrieval-stage analysis, we additionally evaluate K ∈{3,5,10} to examine ground-truth intent coverage at different candidate-set sizes. We additionally report supervised reference results from task-specific training, including a Conformer-Transformer system [21] implemented with NVIDIA NeMo and LLM SFT results using Qwen3-4B [18] and Gemma-3-4B [19].

APEX and the query-based retrieval baseline use the same sentence encoder, backbone LLM, and downstream prediction procedure. Their primary difference is the retrieval scope: APEX applies intent-level candidate selection before exemplar retrieval, whereas the query-based baseline retrieves directly from the complete training set. The zero-shot setting removes contextual demonstrations while retaining the same backbone LLMs. These settings provide a controlled basis for comparing the retrieval strategies without introducing task-specific LLM fine-tuning.

### 4.3. Evaluation metrics

Intent Accuracy and Exact Match Accuracy (EMA) are used to evaluate final SLU performance. Intent Accuracy measures the proportion of test utterances for which the predicted intent matches the reference intent. EMA is used as the primary joint metric and counts a prediction as correct only when the complete semantic output, including both intent and slot information, exactly matches the reference. EMA therefore provides a strict joint measure of intent detection and slot filling performance. We report both metrics on the full test set and on the Low-PPL and High-PPL subsets.

For retrieval-stage analysis, ground-truth intent Coverage@K is reported for $K \in \{3,5,10\}$. Coverage@K measures the proportion of test queries for which the ground-truth intent is covered by the retrieved candidates. For query-based retrieval, coverage is computed over the intents associated with the Top- $K$ retrieved exemplars, whereas for prototype-guided retrieval, it is computed directly over the Top-$K$ retrieved intent prototypes. Coverage@K is used only as a retrieval-stage diagnostic; higher coverage indicates that the correct intent is more frequently retained during retrieval, but does not by itself imply a correct downstream semantic prediction. The number of distinct intents represented by the Top-K query-based exemplars may be smaller than K when multiple exemplars correspond to the same intent. Consequently, Coverage@K is not a strictly matched comparison of intent-selection quality between the two retrieval strategies, since the number of distinct candidate intents may differ. The observed coverage gap may therefore reflect both prototype-guided candidate selection and differences in intent diversity among the retrieved candidates.

Table 1. SLU performance (%) under full-transcript and PPL-stratified conditions. Fine-tuning-free results use 2,974 utterances for All and 2,952 valid-PPL utterances for the PPL subsets; supervised results follow their respective evaluation settings.

| Model | Strategy | All | | Low-PPL (84.1%) | | High-PPL (15.9%) | |
|---|---|---|---|---|---|---|---|
| | | Int. | EMA | Int. | EMA | Int. | EMA |
| ***Supervised Baselines*** | | | | | | | |
| Conformer-Transformer | NVIDIA NeMo E2E | 90.14 | **73.18** | **93.67** | **77.56** | 84.76 | 65.67 |
| Gemma-3-4B | LLM SFT | 90.05 | 72.80 | 92.30 | 75.66 | 80.64 | 58.72 |
| Qwen3-4B | LLM SFT | 89.31 | 72.34 | 91.86 | 75.42 | 77.87 | 57.45 |
| ***Fine-tuning-free In-Context Learning*** | | | | | | | |
| Qwen3-4B | APEX | 86.92 | **56.56** | 89.85 | **60.15** | 75.74 | 39.15 |
| | Query-based Retrieval | 83.69 | 52.22 | 87.91 | 56.04 | 63.83 | 31.49 |
| | Zero-shot | 56.83 | 18.37 | 60.64 | 19.68 | 38.94 | 8.29 |
| Gemma-3-4B | APEX | **87.39** | 55.31 | **89.93** | 57.86 | **76.81** | **43.19** |
| | Query-based Retrieval | 86.84 | 51.34 | 88.09 | 53.38 | 72.66 | 41.06 |
| | Zero-shot | 29.59 | 12.78 | 29.81 | 13.84 | 27.87 | 10.48 |

# 5. RESULTS AND ANALYSIS

## 5.1. Overall SLU performance

Table 1 summarizes SLU performance under the full-transcript and PPL-stratified conditions. The supervised systems are included as reference results for task-specific training and are not intended as like-for-like comparisons with the fine-tuning-free methods considered in this work. Among the fine-tuning-free settings, APEX obtains higher Intent Accuracy and EMA than query-based retrieval with both backbone LLMs. For Qwen3-4B, Intent Accuracy increases from 83.69% to 86.92%, while EMA increases from 52.22% to 56.56%. With Gemma-3-4B, the corresponding scores increase from 86.84% to 87.39% and from 51.34% to 55.31%, respectively. These results correspond to absolute EMA gains of 4.34 and 3.97 percentage points for Qwen3-4B and Gemma-3-4B, respectively.

Both retrieval-based settings also achieve higher scores than zero-shot inference, suggesting that contextual demonstrations can provide useful task-specific information without additional LLM fine-tuning. Compared with direct query-based retrieval, the improvements observed with APEX suggest that intent-level candidate selection can complement example-level-similarity during demonstration retrieval. These results are consistent across both backbone LLMs under the full-transcript condition, although the magnitude of the improvement varies between models.

## 5.2. Impact of utterance predictability

The PPL-stratified results in Table 1 provide a complementary view of performance under different levels of utterance predictability. All evaluated systems show lower EMA on the High-PPL subset than on the Low-PPL subset, indicating an association between lower utterance predictability and increased SLU difficulty under the present evaluation setting. Within the fine-tuning-free settings, APEX maintains higher EMA than query-based retrieval for both backbone LLMs and under both PPL conditions.

For Qwen3-4B, APEX improves EMA over query-based retrieval by 4.11 percentage points on the Low-PPL subset and 7.66 percentage points on the High-PPL subset. For Gemma-3-4B, the corresponding improvements are 4.48 and 2.13 percentage points. The magnitude of these gains differs across backbone models, with a larger High-PPL improvement observed for Qwen3-4B than for Gemma-3-4B. Rather than indicating a uniform effect across models, these results suggest that the benefit of prototype-guided candidate selection may depend partly on the downstream LLM and the characteristics of the input utterances.

## 5.3. Retrieval analysis under different input conditions

Beyond the downstream SLU results in Table 1, Fig. 2 examines retrieval behavior using ground-truth intent Coverage@K. Across the reference-transcript and PPL-stratified conditions, APEX achieves higher ground-truth intent Coverage@K than query-based retrieval for all evaluated values of K. However, as discussed in Section 4.3, the two retrieval strategies differ in the number of distinct intents represented among their Top-K candidates. Therefore, the observed coverage advantage should be interpreted as a property of the overall retrieval strategies rather than as evidence that prototype-based intent ranking alone is superior.

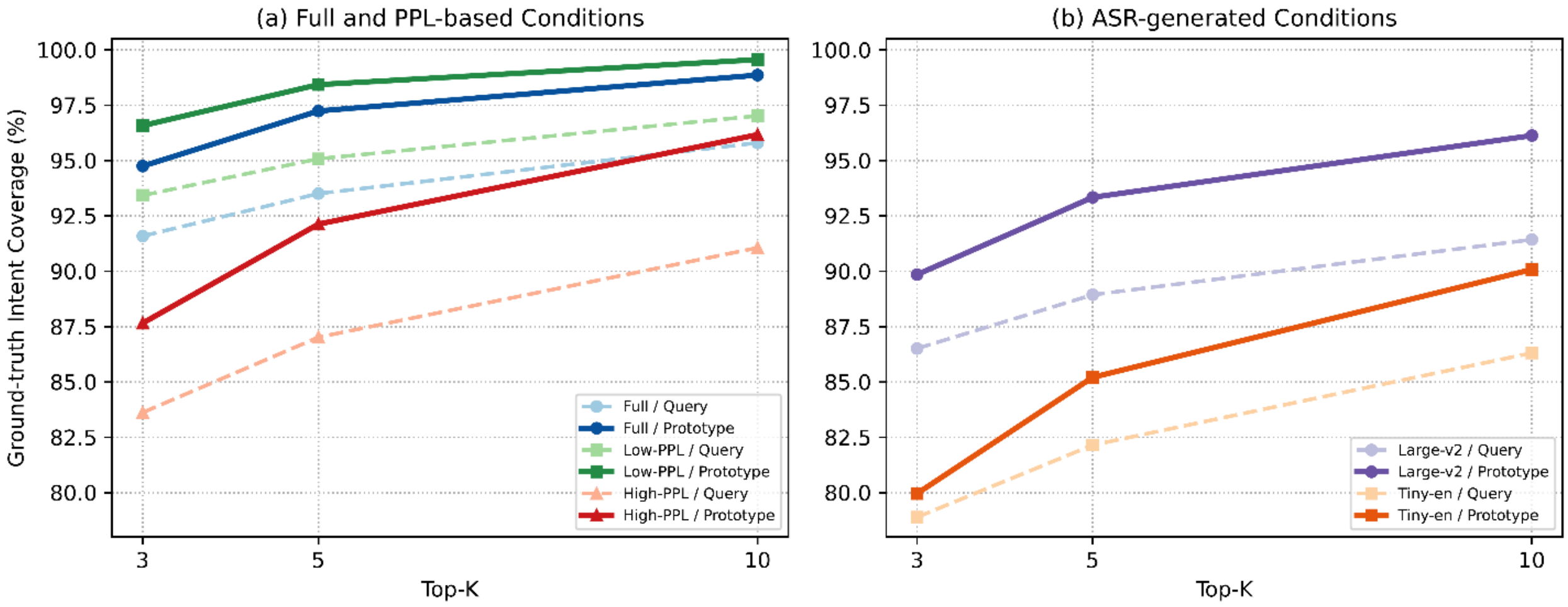

Fig. 2. Ground-truth intent coverage of APEX and query-based retrieval under different input conditions.

A similar pattern is observed with transcripts generated by Whisper Large-v2 and Whisper Tiny-en. Under both ASR-generated conditions, APEX achieves higher ground-truth intent Coverage@K than query-based retrieval across all evaluated values of K. However, this comparison is subject to the same candidate-diversity limitation discussed in Section 4.3. These observations are limited to retrieval-stage coverage and do not establish improvements in final downstream SLU performance under ASR-generated inputs.

## 6. CONCLUSION

This work examines class-level intent information as a lightweight retrieval prior for fine-tuning-free SLU. APEX uses intent prototypes for candidate selection before retrieving query-specific contextual exemplars. Experiments on SLURP show that APEX improves EMA over query-based retrieval by 4.34 and 3.97 percentage points with Qwen3-4B and Gemma-3-4B, respectively, while also showing improvements across both PPL-stratified conditions. Retrieval analysis further shows consistently higher ground-truth intent coverage across the evaluated candidate-set sizes and under ASR-generated conditions. These results suggest that class-level semantic information can complement example-level similarity. The present evaluation is limited to the SLURP dataset and two 4B backbone LLMs, with downstream SLU experiments using a fixed candidate-intent size of K = 5. Moreover, the retrieval-stage coverage comparisons do not isolate the effects of intent ranking from differences in candidate-intent diversity, and downstream SLU performance under ASR-generated transcripts remains unexamined. Future work will investigate additional datasets and backbone models, adaptive retrieval configurations, and downstream SLU performance under ASR-generated inputs.